\documentclass[10pt,twocolumn,letterpaper]{article}

\usepackage{cvpr}
\usepackage{times}
\usepackage{epsfig}
\usepackage{multirow,tabularx}
\usepackage{bm}
\usepackage{graphicx}
\usepackage{amsmath}
\usepackage{amssymb}
\usepackage{breqn}
\usepackage{mathtools}
\usepackage{soul}
\usepackage[pagebackref=true,breaklinks=true,letterpaper=true,colorlinks,bookmarks=false]{hyperref}

\usepackage[breaklinks=true,bookmarks=false]{hyperref}

\cvprfinalcopy 

\def\cvprPaperID{6967} 

\ifcvprfinal\fi
\begin{document}

\title{Heterogeneous Knowledge Distillation using Information Flow Modeling}

\author{N. Passalis, M. Tzelepi and A. Tefas\\
Department of Informatics, Aristotle University of Thessaloniki, Greece\\
{\tt\small \{passalis, mtzelepi, tefas\}@csd.auth.gr}
}

\maketitle
\thispagestyle{empty}

\begin{abstract}
Knowledge Distillation (KD) methods are capable of transferring the knowledge encoded in a large and complex  teacher into a smaller and faster student. Early methods were usually limited to transferring the knowledge only between  the last layers of the networks, while latter approaches were capable of performing multi-layer KD, further increasing the accuracy of the student.
However, despite their improved performance, these methods  still suffer from several limitations that restrict both their efficiency and flexibility. First, existing KD methods typically ignore that neural networks undergo through different learning phases during the training process, which often requires different types of supervision for each one. Furthermore, existing multi-layer KD methods are usually unable to effectively handle networks with significantly different architectures (heterogeneous KD). In this paper we propose a novel KD method that works by modeling the information flow through the various layers of the teacher model and then train a student model to mimic this information flow. The proposed method is capable of overcoming the aforementioned limitations by using an appropriate supervision scheme during the different phases of the training process, as well as by designing and training an appropriate auxiliary teacher model that  acts as a proxy model capable of ``explaining'' the way the teacher works to the student. The effectiveness of the proposed method is demonstrated using four  image datasets and several different evaluation setups.
\end{abstract}

\section{Introduction}

\begin{figure}
	\begin{center}
		\includegraphics[width=0.99\linewidth]{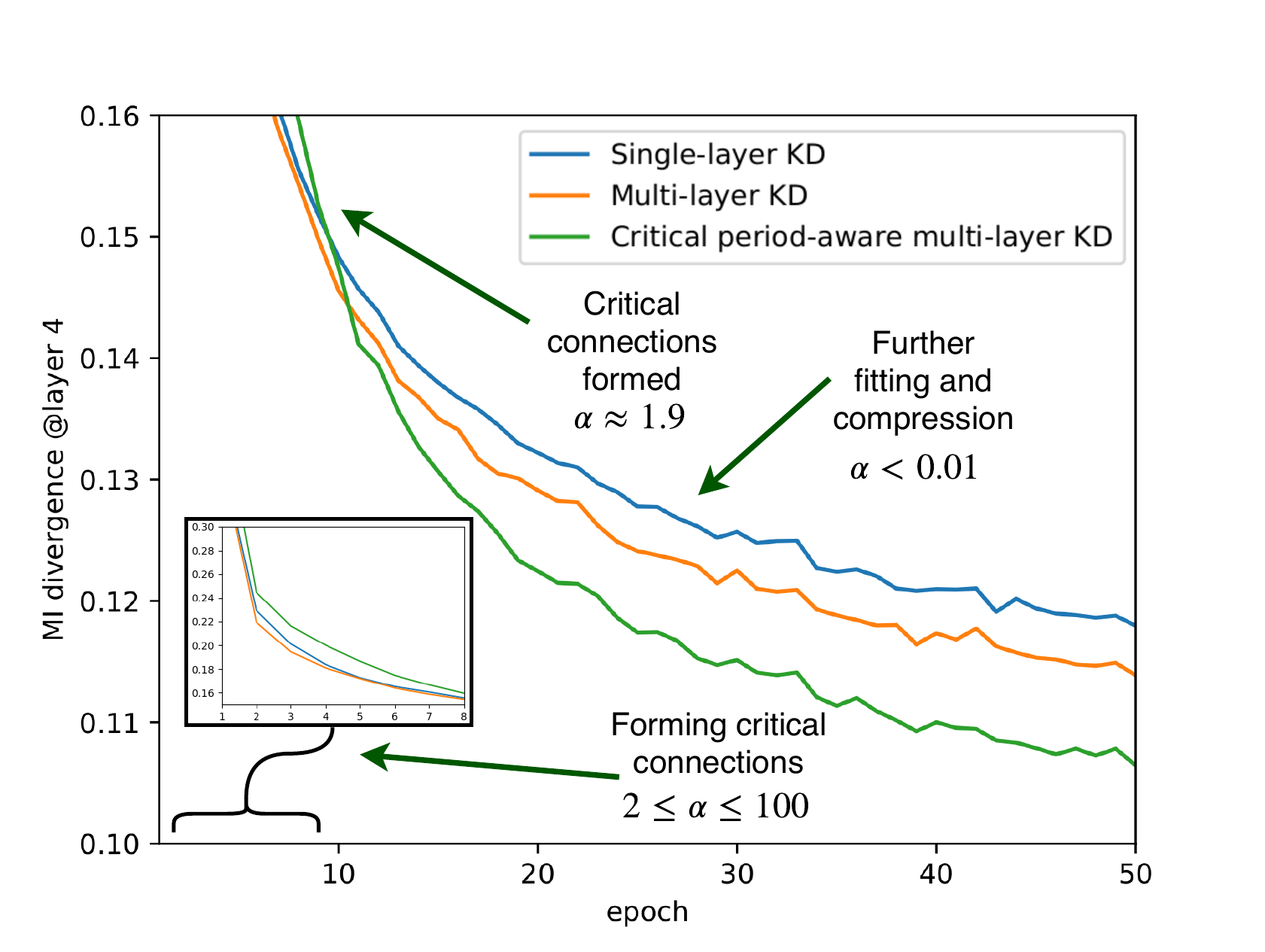}
		\vspace{-1em}
	\end{center}
	\caption{Existing knowledge distillation approaches ignore the existence of critical learning periods when transferring the knowledge, even when multi-layer transfer approaches are used. However, as argued in~\cite{achille2017critical}, the \textit{information plasticity} rapidly declines after the first few training epochs, reducing the effectiveness of knowledge distillation. On the other hand, the proposed method models the information flow in the teacher network and provides the appropriate supervision during the first few critical learning epochs in order to ensure that the necessary connections between successive layers of the networks will be formed. Note that even though this process initially slows down the convergence of the network slightly (epochs 1-8), it allows for rapidly increasing the rate of convergence after the critical learning period ends (epochs 10-25). The parameter $\alpha$ controls the relative importance of transferring the knowledge from the intermediate layers during the various learning phases, as described in detail in Section~\ref{section:proposed}.
	}
	\label{fig:ablation-critical}
\end{figure}

Despite the tremendous  success of Deep Learning (DL) in a wide range of domain~\cite{lecun2015deep}, most DL methods suffer from a significant drawback: powerful hardware is needed for training and deploying DL models. This significantly hinders DL applications on resource-scarce environments, such as embedded and mobile devices, leading to the development of various methods for overcoming these limitations. Among the most prominent methods for this task is \textit{knowledge distillation} (KD)~\cite{hinton2015distilling}, which is also known as \textit{knowledge transfer} (KT)~\cite{yim2017gift}. These approaches aim to transfer the knowledge encoded in a large and complex neural network into a smaller and faster one. In this way, it is possible to increase the accuracy of the smaller model, compared to the same model trained without employing KD. Typically, the smaller model is called \textit{student} model, while the larger model is called \textit{teacher} model.

Early KD approaches focused on transferring the knowledge between the last layer of the teacher and student models~\cite{compression-model,	hinton2015distilling, passalis2018learning,tang2016recurrent,tzeng2015simultaneous,yu2019learning}. This allowed for providing richer training targets to the student model, which capture more information regarding the similarities between different samples, reducing overfitting and increasing the student's accuracy. Later methods further increased the efficiency of KD by modeling and transferring the knowledge encoded in the \textit{intermediate} layers of the teacher~\cite{romero2014fitnets, yim2017gift, zagoruyko2016paying}. These approaches usually attempt to implicitly model the way information gets transformed through the various layers of a network, providing additional \textit{hints} to the student model regarding the way that the teacher model process the information.

Even though these methods were indeed able to further increase the accuracy of models trained with KD, they also suffer from several limitations that restrict both their efficiency and flexibility. First, note that neural networks exhibit an evolving behavior, undergoing several different and distinct phases during the training process. For example, during the first few epochs critical connections are formed~\cite{achille2017critical}, defining almost permanently the future \textit{information flow} paths on a network. After fixing these paths, the training process can only fine-tune them, while forming new paths is significantly less probable after the critical learning period ends~\cite{achille2017critical}. After forming these critical connections, the fitting and compression (when applicable) phases follow~\cite{ shwartz2017opening, saxe2018information}. Despite this dynamic time-dependent behavior of neural networks, virtually all existing KD approaches ignores the phases that neural networks undergo during the training. This observation leads us to the first research question of this paper: \textit{Is a different type of supervision needed during the different learning phases of the student and is it possible to use a stronger teacher  to provide this supervision?}

To this end, we propose a simple, yet effective way to exploit KD to train a student that mimics the information flow paths of the teacher, while also providing further evidence confirming the existence of critical learning periods during the training phase of a neural network, as originally described in~\cite{achille2017critical}. Indeed, as also demonstrated in the ablation study shown in Fig.~\ref{fig:ablation-critical}, providing the correct supervision during the critical learning period of a neural network can have a significant effect on the overall training process, increasing the accuracy of the student model. More information regarding this ablation study are provided in Section~\ref{section:experiments}. It is worth noting that the additional supervision, which is employed to ensure that the student will form similar information paths to the teacher, actually slows down the learning process until the critical learning period is completed. However, after the information flow paths are formed, the rate of convergence is significantly accelerated compared to the student networks that do not take into account the existence of critical learning periods.

\begin{figure}
	\begin{center}
		\includegraphics[width=0.99\linewidth]{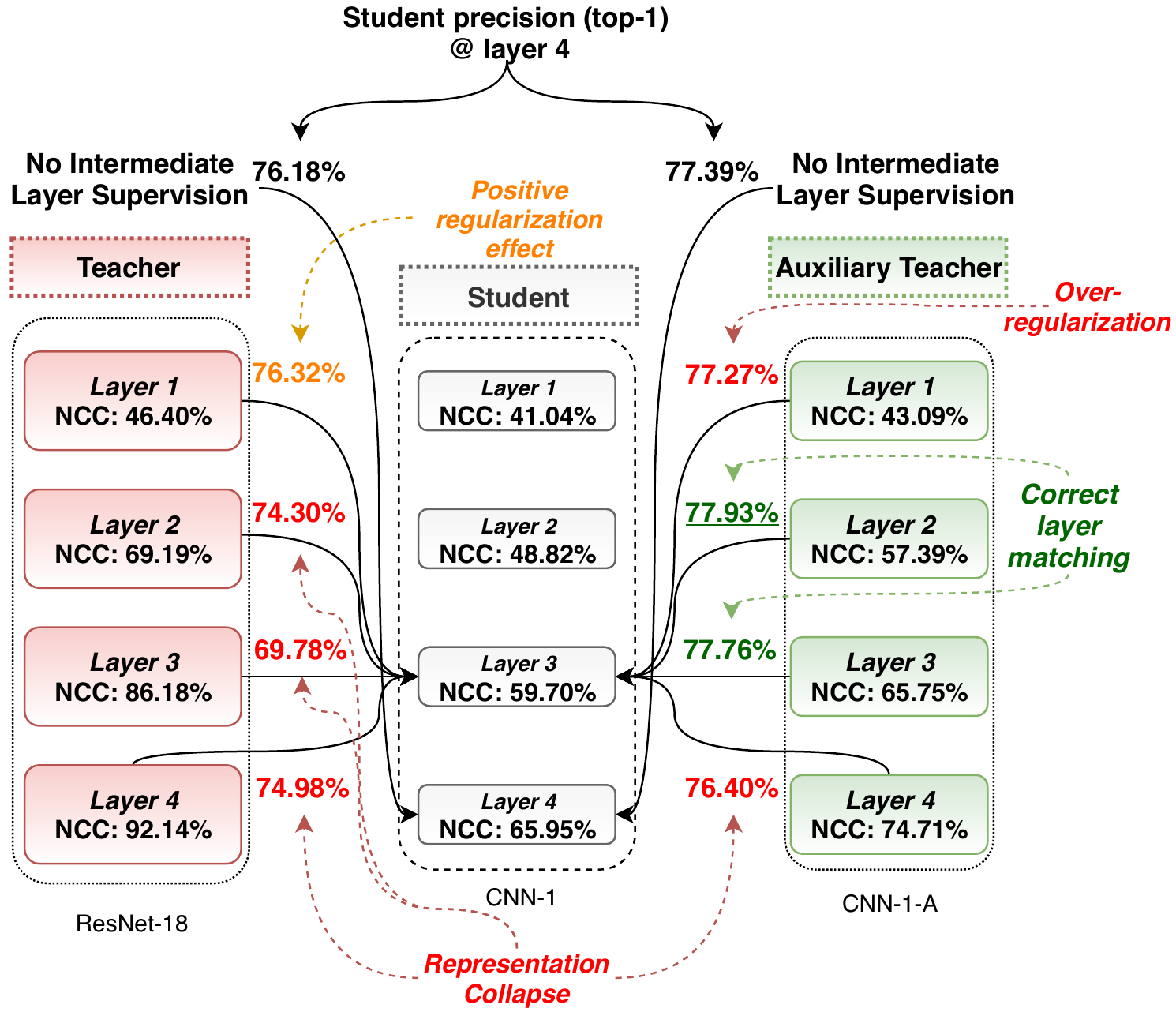}
		\end{center}
	
	\caption{Examining the effect of transferring the knowledge from different layers of a teacher model into the third layer of the student model. Two different teachers are used, a strong teacher (ResNet-18, where each layer refers to each layer block) and an auxiliary teacher (CNN-1-A). The nearest centroid classifier accuracy (NCC) is reported for the representations extracted from each layer (in order to provide an intuitive measure of how each layer transforms the representations extracted from the input data). The final precision is reported for a student model trained by either not using intermediate layer supervision (upper black values) or by using different layers of the teacher (4 subsequent precision values). Several different phenomena are observed when the knowledge is transferred from different layers, while the proposed auxiliary teacher allows for achieving the highest precision and provides a straightforward way to match the layers between the models (the auxiliary teacher transforms the data representations in a way that is closer to the student model, as measure through the NCC accuracy).
	}
		\vspace{-1em}
	\label{fig:ablation-layers}
\end{figure}

Another limitation of existing KD approaches that employ multiple intermediate layers is their ability to handle \textit{heterogeneous multi-layer knowledge distillation}, i.e., transfer the knowledge between teachers and students with vastly different architectures. 
Existing methods almost exclusively  use network architectures that provide a trivial one-to-one matching between the layers of the student and teacher, e.g., ResNets with the same number of blocks are often used, altering only the number of layers inside of each residual block~\cite{yim2017gift, zagoruyko2016paying}. Many of these approaches, such as~\cite{yim2017gift}, are even more restrictive, also requiring the layers of the teacher and student to have the same dimensionality. As a result, it is especially difficult to perform multi-layer KD between networks with vastly different architectures, since even if just one layer of the teacher model is incorrectly matched to a layer of the student model, then the accuracy of the student can be significantly reduced, either due to over-regularizing the network or by forcing to early compress the representations of the student. This behavior is demonstrated in Fig.~\ref{fig:ablation-layers}, where the knowledge is transferred from the 3rd layer of two different teachers to various layers of the student. These findings lead us to the second research question of this paper: \textit{Is it possible to handle heterogeneous KD in a structured way to avoid such phenomena? }

To this end, in this work, we propose a simple, yet effective approach for training an \textit{auxiliary teacher} model, which is closer to the architecture of the student model. This auxiliary teacher is responsible for \textit{explaining} the way the larger teacher works to the student model. Indeed, this approach can significantly increase the accuracy of the teacher, as demonstrated both in Fig~\ref{fig:ablation-layers}, as well as on the rest of the experiments conducted in this paper.  It is worth noting that during our initial experiments it was almost impossible to find a layer matching that would actually help us to improve the accuracy of the student model without first designing an appropriate auxiliary teacher model, highlighting the importance of using auxiliary teachers in heterogeneous KD scenarios, as also highlighted in~\cite{mirzadeh2019improved}.

The main contribution of this paper is proposing a KD method that works by modeling the information flow through the teacher model and then training a student model to mimic this information flow. However, as it was explained previously and experimentally demonstrated in this paper, this process is often very difficult, especially when there is no obvious layer matching between the teacher and student models, which can often process the information in vastly different ways. In fact, even a single layer mismatch, i.e., overly regularizing the network or forcing for an early compression of the representation, can significantly reduce the accuracy of the student model. To overcome these limitations the proposed method works by a) designing and training an appropriate auxiliary teacher model that  allows for a direct and effective one-to-one  matching between the layers of the student and teacher models, as well as b) employing a critical-learning aware KD scheme that ensures that critical connections will be formed allowing for effectively mimicking the teacher's information flow instead of just learning a student that mimics the output of the student.

The effectiveness of the proposed method is demonstrated using several different tasks, ranging from metric learning and classification to mimicking handcrafted feature extractors for providing fast neural network-based implementations for low power embedded hardware.  The experimental evaluation also includes an extensive representation learning evaluation, given its  increasing importance in many embedded DL and robotic applications and following the evaluation protocol of recently proposed KD methods~\cite{passalis2018learning,yu2019learning}. 
An open-source implementation of the proposed method is provided in \url{https://github.com/passalis/pkth}.

The rest of the paper is structured as follows. First, the related work is briefly discussed and compared to the proposed method in Section~\ref{section:related-work}. Then, the proposed method is presented in Section~\ref{section:proposed}, while the experimental evaluation is provided in Section~\ref{section:experiments}. Finally, conclusions are drawn in Section~\ref{section:conclusions}.

\section{Related Work}
\label{section:related-work}

A large number of knowledge  transfer methods which build upon the  neural network distillation approach have been proposed~\cite{ahn2019variational,compression-model,hinton2015distilling,tang2016recurrent,tzeng2015simultaneous}. These methods typically use a teacher model to generate soft-labels and then use these soft-labels for training a smaller student network. It is worth noting that several extensions to this approach have  been proposed. For example, soft-labels can be used for pre-training a large network~\cite{tang2015knowledge} and performing domain adaption~\cite{tzeng2015simultaneous}, while an embedding-based approach  for transferring the knowledge was proposed in~\cite{passalis2018unsupervised}.  Also, {online} distillation methods, such as~\cite{anil2018large, zhang2018deep}, employ a co-training strategy, training both the student and teacher models simultaneously. However, none of these approaches take into account that deep neural networks transition through several learning phases, each one with different characteristics, which requires handling them in different ways. On the other hand, the proposed method models the information flow in the teacher model and then employs a weighting scheme that provides the appropriate supervision during the initial critical learning period of the student, ensuring that the critical connections and information paths formed in the teacher model will be transferred to the student.

Furthermore, several methods that support multi-layer KD have been proposed, such as using hints~\cite{romero2014fitnets}, the flow of solution procedure matrix (FSP)~\cite{yim2017gift}, attention transfer~\cite{zagoruyko2016paying}, or singular value decomposition to extract major features from each layer~\cite{lee2018self}.  However, these approaches  usually only target networks with compatible architecture, e.g., residual networks with the same number of residual blocks, both for the teacher and student models. Also, it is not straightforward to use them to successfully transfer the knowledge between heterogeneous models, since even a slight layer mismatch can have a devastating effect on the student's accuracy, as demonstrated in Fig.~\ref{fig:ablation-layers}. It is also worth noting that we were actually unable to effectively apply most of these methods for heterogenous KD, since either they do not support transferring the knowledge between layers of different dimensionality, e.g.,~\cite{yim2017gift}, or they are prone to over-regularization or representation collapse (as demonstrated in Fig.~\ref{fig:ablation-layers}) reducing the overall performance of the student.  

In contrast with the aforementioned approaches, the proposed method provides a way to perform heterogeneous multi-layer KD by appropriately designing and training an auxiliary network and exploiting the knowledge encoded by the earlier layers of this network.  In this way, the proposed method provides an efficient way for handling any possible network architecture by employing an auxiliary network that is close to the architecture of the student model regardless the architecture of the teacher model. Using the proposed auxiliary network strategy ensures that the teacher model will transform the representations extracted from the data in a way compatible with the student model, allowing for providing a one-to-one matching between the intermediate layers of the networks. It is also  worth nothing that the use of a similar auxiliary network, which is used as an intermediate step for KD, was also proposed in~\cite{mirzadeh2019improved}. However in contrast with the proposed method, the auxiliary network used in~\cite{mirzadeh2019improved} was employed for merely improving the performance of KD between the final classification layers, instead of designing an auxiliary network that can facilitate efficient multi-layer KD, as proposed in this paper. Finally, to the best of our knowledge, in this work we propose the first architecture-agnostic probabilistic KD approach that works by modeling the information flow through the various layers of the teacher model using a hybrid kernel formulation, can support heterogeneous network architectures and can effectively supervise the student model during its critical learning period.

\section{Proposed Method}
\label{section:proposed}

Let $\mathcal{T} =\{ \mathbf{t}_1, \mathbf{t}_2, \dots, \mathbf{t}_N \}$ denote the \textit{transfer} set that contains $N$ transfer samples and it is  used to transfer the knowledge from the teacher model to the student model. Note that the proposed method can also work in a purely unsupervised fashion and, as a result, unlabeled data samples can be also used for transferring the knowledge. Also, let $\mathbf{x}_i^{(l)} =f(\mathbf{t}_i, l)$ denote the representation extracted from the $l$-th layer of the teacher model $f(\cdot)$ and  $\mathbf{y}_i^{(l)}=g(\mathbf{t}_i, l, \mathbf{W})$  denote the representation extracted from the $l$-th layer of the student model  $g(\cdot)$. Note that the trainable parameters of the student model are denoted by $\mathbf{W}$. The proposed method aims to train the student model  $g(\cdot)$, i.e., learn the appropriate parameters $\mathbf{W}$, in order to  ``mimic'' the behavior of $f(\cdot)$ as close as possible.  

Furthermore, let $\mathcal{X}^{(l)}$ denote the random variable that describes the representation extracted from the $l$-th layer of the teacher model and $\mathcal{Y}^{(l)}$ the corresponding random variable for the student model. Also, let $\mathcal{Z}$ denote the random variable that describes the training targets for the teacher model. In this work, the {information flow} of the teacher network is defined as progression of mutual information between every layer representation of the network and the training targets, i.e.,  $I(\mathcal{X}^{(l)},  \mathcal{Z}) \ \forall \  l$.  Note that even though the training targets are required for modeling the information flow, they are not actually needed during the KD process, as we will demonstrate later. Then, we can define the information flow vector that characterizes the way the network process information as:
\begin{eqnarray}
\bm{\omega}_t \coloneqq \left[I(\mathcal{X}^{(1)},  \mathcal{Z}),  \dots, I(\mathcal{X}^{(N_{L_t})}, \mathcal{Z})\right]^T \in \mathbb{R}^{N_{L_t}},
\end{eqnarray}
where $N_{L_t}$ is the number of layers of the teacher model. Similarly, the information flow vector for the student model is defined as:
\begin{eqnarray}
\bm{\omega}_s \coloneqq \left[I(\mathcal{Y}^{(1)},  \mathcal{Z}),  \dots, I(\mathcal{Y}^{(N_{L_s})}, \mathcal{Z})\right]^T \in \mathbb{R}^{N_{L_s}},
\end{eqnarray}
where again $N_{L_s}$ is the number of layers of the student model. The proposed method works by minimizing the divergence between the information flow in the teacher and student models, i.e., $D_{F} (\bm{\omega}_s, \bm{\omega}_t)$, where $D_{F}(\cdot)$ is a metric used to measure the divergence between two, possibly heterogeneous, networks.  To this end, the information flow divergence is defined as the sum of squared differences between each paired element of the information flow vectors:
\begin{dmath}
	\label{eq:flow-divergence}
	D_{F} (\bm{\omega}_s, \bm{\omega}_t) = \sum_{i=1}^{N_{L_s}} \left( [\bm{\omega}_s]_i - [\bm{\omega}_t]_{\kappa(i)}   \right)^2,
\end{dmath}
where the layer of the teacher $\kappa(i)$ is chosen in order to minimize the divergence with the corresponding layer of the teacher:
\begin{equation}
\label{eq:matchinglayer}
\kappa(i) = 
\begin{cases}
N_{L_t} & \text{if $i = N_{L_s} $}\\
\arg \min _{j}  ([\bm{\omega}_s]_i- [\bm{\omega}_t]_{j})^2, & \text{otherwise}
\end{cases}
\end{equation}
and the notation $[\mathbf{x}]_i$ is used to refer to the $i$-th element of vector $\mathbf{x}$. This definition employs the optimal matching between the layers (considering the discriminative power of each layer), except from the final one, which corresponds to the task hand.  In this way, it allows for measuring the flow divergence between networks with different architectures. At the same time it is also expected to minimize the impact of  over-regularization and/or representation collapse phenomena, such as those demonstrated in Fig.~\ref{fig:ablation-layers}, which often occur when there is large divergence between the layers used for transferring the knowledge. However, this also implies that for networks with vastly different architectures or for networks not yet trained for the task at hand, the same layer of the teacher may be used for transferring the knowledge to multiple layers of the student model, leading to a significant loss of granularity during the KD and leading to stability issues. In Subsection~\ref{sub:2} we provide a simple, yet effective way to overcome this issue by using auxiliary teacher models. Note that more advanced methods, such as employing fuzzy assignments between different sets of layers can be also used.

\subsection{Tractable Information Flow Divergence Measures using Quadratic Mutual Information}
In order to effectively transfer the knowledge between two different networks, we have to provide an efficient way to calculate the mutual information, as well as to train the student model to match the mutual information between  two layers of different networks. Recently, it has been demonstrated that when the Quadratic Mutual Information (QMI)~\cite{torkkola2003feature} is used, it is possible to efficiently minimize the difference between the mutual information of a specific layer of the teacher and student by appropriately relaxing the optimization problem~\cite{passalis2018learning}. More specifically, the problem of matching the mutual information between two layers can be reduced into a simpler probability matching problem that involves only the pairwise interactions between the transfer samples. Therefore, to transfer the knowledge between a specific layer of the student and an another layer of the teacher, it is adequate to minimize the divergence between the teacher's and student's conditional probability distributions, which can be estimated as~\cite{passalis2018learning}:
\small
\begin{equation} 
\label{eq:teacher-prob} 
p^{(t, l_t)}_{i|j} = \frac{K(\mathbf{x}^{(l_t)}_{i}, \mathbf{x}^{(l_t)}_{j})}{\sum_{i=1, i\neq j}^N K(\mathbf{x}^{(l_t)}_{i}, \mathbf{x}^{(l_t)}_{j})} \in [0,1], 
\end{equation} 
\normalsize
and
\small
\begin{equation} 
p^{(s, l_s)}_{i|j} = \frac{K(\mathbf{y}^{(l_s)}_{i}, \mathbf{y}^{(l_s)}_{j})}{\sum_{i=1, i\neq j}^N K(\mathbf{y}^{(l_s)}_{i}, \mathbf{y}^{(l_s)}_{j})} \in [0,1],
\end{equation} 
\normalsize
where $K(\cdot)$ is a kernel function and $l_t$ and  $l_s$ refer to the student and teacher layers used for the transfer. These probabilities also express how probable is for each sample to select each of its neighbors~\cite{maaten2008visualizing}, modeling in this way the geometry of the feature space, while matching these two distributions also ensures that the mutual information between the models and a set of (possibly unknown) classes is maintained~\cite{passalis2018learning}. Note that the actual training labels are not required during this process, and, as a result, the proposed method can work in a purely unsupervised fashion.

The kernel choice can have a significant effect on the quality of the KD, since it alters how the mutual information is estimated~\cite{passalis2018learning}. Apart from the well known Gaussian kernel, which is however often hard to tune, other kernel choices include cosine-based kernels~\cite{passalis2018learning}, e.g.,
$K_c(\mathbf{a}, \mathbf{b}) = \frac{1}{2}(\frac{\mathbf{a}^T\mathbf{b}}{||\mathbf{a}||_2 ||\mathbf{b}||_2} + 1)$, where $\mathbf{a}$ and $\mathbf{b}$ are two vectors,
and the T-student kernel, i.e.,
$K_T(\mathbf{a}, \mathbf{b}) = \frac{1}{1+||\mathbf{a}-\mathbf{b}||_2^d}$, where $d$ is typically set to 1. Selecting the most appropriate kernel for the task at hand can lead to significant performance improvements, e.g., cosine-based kernels perform better for retrieval tasks, while using kernel ensembles, i.e., estimating the probability distribution using multiple kernels, can also improve the robustness of mutual information estimation. Therefore, in this paper a hybrid objective that aims at minimizing the divergence calculated using both the cosine kernel, which ensures the good performance of the learned representation for retrieval tasks, and the T-student kernel, which experimentally demonstrated good performance for classification tasks, is used:
\begin{equation}
\label{eq:loss2}
\mathcal{L}^{(l_t, l_s)} = \mathcal{D}(\mathcal{P}_c^{(t, l_t)}|| \mathcal{P}_{c}^{(s, l_s)}) + \mathcal{D}(\mathcal{P}_T^{(t, l_t)}|| \mathcal{P}_{T}^{(s, l_s)}), 
\end{equation}
where $\mathcal{D}(\cdot)$ is a probability divergence metric and the notation $\mathcal{P}_c^{(t, l_t)}$  and $\mathcal{P}_T^{(t, l_t)}$ is used to denote the conditional probabilities of the teacher calculated using the cosine and T-student kernels respectively. Again, the representations used for KD were extracted from the $l_t$-th/$l_s$-th layer. The student probability distribution is denoted similarly by $\mathcal{P}_c^{(s, l_s)}$  and $\mathcal{P}_T^{(s, l_s)}$. The divergence between these distributions can be calculated using a symmetric version of the Kullback-Leibler (KL) divergence, the  Jeffreys divergence~\cite{jeffreys1946invariant}:
\small
\begin{dmath}
	\mathcal{D}(\mathcal{P}^{(t, l_t)}||\mathcal{P}^{(s, l_s)})=\\ \sum_{i=1}^{N} \sum_{j=1, i\neq j}^{N} \left( {p^{(t, l_t)}_{j|i}} - {p^{(s, l_s)}_{j|i}} \right) \cdot \left( \log {p^{(t, l_t)}_{j|i}} - \log{p^{(s, l_s)}_{j|i}}\right),
\end{dmath}
\normalsize
which can be sampled at a finite number of points during the optimization, e.g., using batches of 64-128 samples. This batch-based strategy has been successfully  employed in a number of different works~\cite{passalis2018learning, yu2019learning}, without any significant effect on the optimization process.

\subsection{Auxiliary Networks and Information Flow}
\label{sub:2}

Even though the flow divergence metric defined in (\ref{eq:flow-divergence}) takes into account the way different networks process the information, it suffers from a significant drawback: if the teacher process the information in a significantly different way compared to the student, then the same layer of the teacher model might be used for transferring the knowledge to multiple layers of the student model, leading to a significant loss in the granularity of information flow used for KD. Furthermore, this problem can also arise even when the student model is capable of processing the information in a way compatible with the teacher, but it has not been yet appropriately trained for the task at hand. To better understand this, note  that the information flow divergence in~(\ref{eq:flow-divergence}) is calculated based on the estimated mutual information and not the actual learning capacity of each model. Therefore, directly using the flow divergence definition presented in~(\ref{eq:flow-divergence}) is not optimal for KD. It is worth noting that this issue is especially critical for every KD method that employs multiple layer, since as we demonstrate in Section~\ref{section:experiments},  if the layer pairs are not carefully selected, the accuracy of the student model is often lower compared to a model trained without using multi-layer transfer at all.

\begin{figure}
	\begin{center}
		\includegraphics[width=0.99\linewidth]{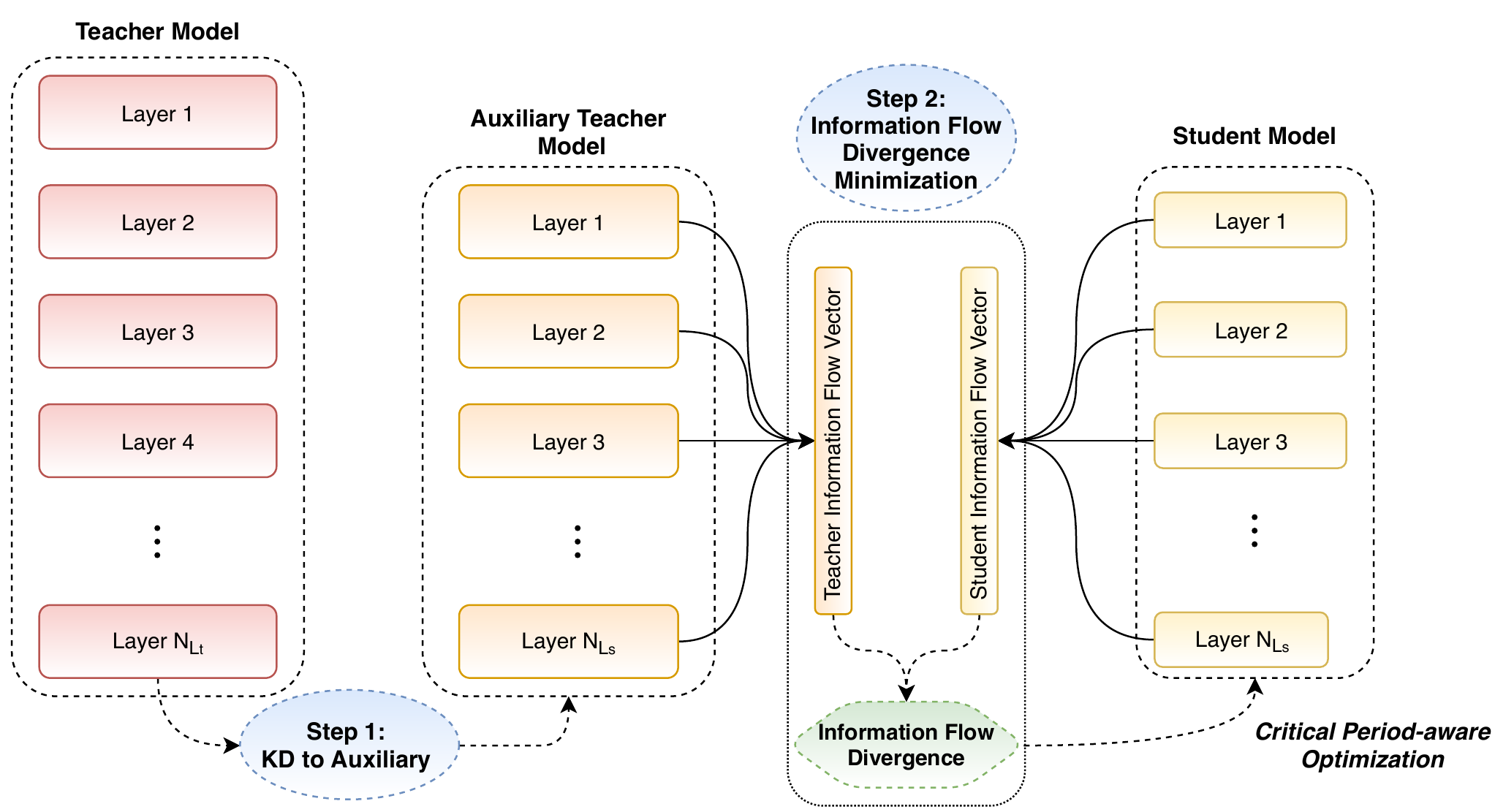}
	\end{center}
	
	\caption{First, the knowledge is transferred to an appropriate auxiliary teacher, which will better facilitate the process of KD. Then, the proposed method minimizes the information flow divergence between the two models, taking into account the existence of critical learning periods.
	}
		\vspace{-1em}
	\label{fig:proposed}
\end{figure}

Unfortunately, due to the poor understanding of the way that neural networks transform the probability distribution of the input data, there is currently no way to  select the most appropriate layers for transferring the knowledge \textit{a priori}. This process can be especially difficult and tedious, especially when the architectures of the student and teacher differ a lot. To overcome this critical limitation in this work we propose constructing an appropriate auxiliary proxy for the  teacher model, that will allow for directly matching between all the layers of the auxiliary model and the student model, as shown in Fig.~\ref{fig:proposed}. In this way, the proposed method employs an \textit{auxiliary} network, that has a compatible architecture with the student model, to better facilitate the process of KD. A simple, yet effective approach for designing the auxiliary network is employed in this work: the auxiliary network follows the same architecture as the student model, but using twice the neurons/convolutional filters per layer. Thus, the greater learning capacity of the auxiliary network ensures that enough knowledge will be always available to the auxiliary network (when compared to the student model), leading to better results compared to directly transferring the knowledge from the teacher model. Designing the most appropriate auxiliary network is an open research area and significantly better ways than the proposed one might exist. However, even this simple approach was adequate to significantly enhance the performance of KD and demonstrate the potential of information flow modeling, as further demonstrated in the ablation studies provided in Section~\ref{section:experiments}. Also, note that a hierarchy of auxiliary teachers can be trained in this fashion, as also proposed in~\cite{mirzadeh2019improved}.

The final loss used be optimize the student model, when an auxiliary network is employed, is calculated as:
\begin{equation}
\mathcal{L} = \sum_{i=1}^{N_{L_s}}  \alpha_{i} \mathcal{L}^{(i, i)}, 
\end{equation}
where $\alpha_{i}$ is a hyper-parameter that controls the relative weight of transferring the knowledge from the $i$-th layer of the teacher to the $i$-th layer of the student and the loss $ \mathcal{L}^{(i, i)}$ defined in (\ref{eq:loss2}) is calculated using the auxiliary teacher, instead of the initial teacher. The value of $\alpha_{i}$ can be dynamically selected during the training process, to ensure that the applied KD scheme takes into account the current learning state of the network, as further discussed in Subsection~\ref{sub:3}. Finally, stochastic gradient descent is employed to train the student model: $
\Delta \mathbf{W} =  - \eta \frac{\partial \mathcal{L}}{ \partial \mathbf{W}}$,  where $\mathbf{W}$ is the matrix with the parameters of the student model and $\eta$ is the employed learning rate.

\subsection{Critical Period-aware Divergence Minimization}
\label{sub:3}

Neural networks transition through different learning phases during the training process, with the first few epochs being especially critical for the later behavior of the network~\cite{achille2017critical}. Using a stronger teacher model provides the opportunity of guiding the student model during the initial critical learning period in order to form the appropriate connectivity between the layers, before the information plasticity declines. However, just minimizing the information flow divergence does not ensure that the appropriate connections will be formed. To better understand this, we have to consider that the gradients back-propagated through the network depend both on the training target, as well as on the initialization of the network. Therefore, for a randomly initialized student, the task of forming the appropriate connections between the intermediate layers might not facilitate the final task at hand (until reaching a certain critical point).
This was clearly demonstrated in Fig~\ref{fig:ablation-critical}, where the convergence of the network was initially slower, when the proposed method was used, until reaching the point at which the critical learning period ends and the convergence of the network is accelerated.

Therefore, in this work we propose using an appropriate weighting scheme for calculating the value of the hyper-parameter $\alpha_i$ during the training process. More specifically, during the critical learning period a significantly higher weight is given to match the information flow for the earlier layers, ignoring the task at hand dictated by the final layer of the teacher, while this weight gradually decays to 0, as the training process progresses. Therefore, the parameter $\alpha_i$ is calculated as:
\begin{dmath}
	\alpha_i = \begin{cases}
		1, & \text{if } i = N_{L_S}\\
		\alpha_{init} \cdot \gamma ^k   & \text{otherwise}
	\end{cases},
\end{dmath}
where $k$ is the current training epoch, $\gamma$ is a decay factor and $\alpha_{init}$ is the initial weight used for matching the information flow in the intermediate layers. The parameter $\gamma$ was set to $0.7$, while $\alpha_{init}$ was set to $100$  for all the experiments conducted in this paper (unless otherwise stated). Therefore, during the first few  epochs  (1-10) the final task at hand has a minimal impact on the optimization objective. However, as the training process progresses the importance of matching the information flow for the intermediate layers gradually diminishes and the optimization switches to fine-tuning the network for the task at hand.

\section{Experimental Evaluation}
\label{section:experiments}

The experimental evaluation of the proposed method is provided in this Section. The proposed method was evaluated using four different datasets (CIFAR-10~\cite{krizhevsky2009learning}, STL-10~\cite{coates2011analysis}, CUB-200~\cite{welinder2010caltech} and SUN Attribute~\cite{patterson2012sun} datasets) and compared to four competitive KD methods: neural network distillation~\cite{hinton2015distilling}, hint-based transfer~\cite{romero2014fitnets}, probabilistic knowledge transfer (PKT)~\cite{passalis2018learning} and metric knowledge transfer (abbreviated as MKT)~\cite{yu2019learning}. A variety of different evaluation setups were used to evaluate various aspects of the proposed method. Please refer to the appendix for a detailed description of the employed networks and evaluation setups.

\begin{table}
	\caption{Metric Learning Evaluation: CIFAR-10}
	\label{table:cifar10-metric}
	\vspace{0.5em}
	\centering
	\scriptsize
	\begin{tabular}{l|cccc}
		\textbf{Method} & \textbf{mAP (e)} &  \textbf{mAP (c)}  & \textbf{top-100 (e)} &  \textbf{top-100 (c)}\\
		
		\hline
		
		\hline
		
		\multicolumn{5}{c}{\textbf{Baseline Models}}\\
		\hline
		Teacher (ResNet-18) & $87.18$ & $90.47$ & $92.15$ & $92.26$ \\
		Aux. (CNN1-A) & $62.12$ & $66.78$ & $73.72$ & $75.91$\\
		\hline
		
		\hline
		\multicolumn{5}{c}{\textbf{With Constrastive Supervision}}\\
		\hline
		Student (CNN1) & $47.69$ & $48.72$ & $57.46$ & $58.50$  \\
		\hline
		Hint.  & $43.56$ & $48.73$ & $\mathbf{60.44}$ & $\mathbf{62.43}$
		\\
		MKT & $45.34$ & $46.84$ & $55.89$ & $57.10$
		\\
		PKT  & $48.87$ & $49.95$ & $58.44$ & $59.48$
		\\
		\hline
		Hint-H & $43.24$ & $47.46$ & $58.97$ & $61.07$  
		\\
		MKT-H &  $44.83$ & $47.12$ & $56.28$ & $57.90$  
		\\
		PKT-H & $48.69$ & $50.09$ & $58.71$ & $60.20$   \\
		Proposed & $\mathbf{49.55}$ & $\mathbf{50.82}$ & ${59.50}$ & $60.79$ \\
		\hline
		
		\hline
		\multicolumn{5}{c}{\textbf{Without Constrastive Supervision}}\\
		\hline
		Student (CNN1) &  $35.30$ & $39.00$ & $55.87$ & $58.77$  \\
		\hline
		Distill. & $37.39$ & $40.53$ & $56.17$ & $58.56$\\
		Hint. & $43.99$ & $48.99$ & $60.69$ & $62.42$ \\
		MKT & $36.26$ & $38.20$ & $50.55$ & $52.72$ \\
		PKT  & $48.07$ & $51.56$ & $60.02$ & $62.50$ \\
		\hline
		
		Hint-H & $42.65$ & $46.46$ & $58.51$ & $60.59$   \\
		MKT-H &  $41.16$ & $43.99$ & $55.10$ & $57.63$  \\
		PKT-H &  $48.05$ & $51.73$ & $60.39$ & $63.01$ \\
		Proposed & ${\mathbf{49.20}}$ & $\underline{\mathbf{53.06}}$ & $\underline{\mathbf{61.54}}$ & $\underline{\mathbf{64.24}}$ \\

	\end{tabular}
	
\end{table}

First, the proposed method was evaluated in a metric learning setup using the CIFAR-10 dataset (Table~\ref{table:cifar10-metric}). The methods were evaluated under two different settings: a) using contrastive supervision (by adding a contrastive loss term in the loss function~\cite{hadsell2006dimensionality}), as well as b) using a purely unsupervised setting (cloning the responses of the powerful teacher model). The simple variants (Hint, MKT, PKT) refer to transferring the knowledge only from the penultimate layer of the teacher, while the ``-H'' variants refer to transferring the knowledge simultaneously from all the layers of the auxiliary model. The abbreviation ``e'' is used to refer to retrieval using the Euclidean similarity metric, while ``c'' is used to refer to retrieval using the cosine similarity.

\begin{table}
	\caption{Classification Evaluation: CIFAR-10}
	\label{table:cifar10-classification}
	\vspace{0.5em}
	\centering
	\small
	\begin{tabular}{l|cccc|c}
		\textbf{Method} & \textbf{Train Accuracy} &  \textbf{Test Accuracy} \\
		
		\hline
		
		\hline
		
		Distill & $72.50$ & $70.68$ \\
		Hint. & $71.29$ & $70.59$  \\
		MKT & $69.73$ & $69.13$ 
		\\
		PKT &  $72.70$ & $70.44$ 
		\\
		\hline
		
		Hint-H & $70.93$ & $69.52$  \\
		MKT-H  & $69.67$ & $68.82$  \\
		PKT-H & $73.43$ & $71.44$ \\
		Proposed & $73.24$ & $\mathbf{71.97}$ \\

	\end{tabular}
	
\end{table}

First, note that using all the layers for distilling the knowledge provides small to no improvements in the retrieval precision, with the exception of the MKT method (when applied without any form of supervision). Actually, in some cases (e.g., when hint-based transfer is employed) the performance when multiple layers are used is worse. This behavior further confirms and highlights the difficulty in applying multi-layer KD methods between heterogeneous architectures. Also, using contrastive supervision seems to provide more consistent results for the competitive methods, especially for the  MKT method. Using the proposed method leads to a significant increase in the mAP, as well as in the top-K precision. For example, mAP (c) increases by over 2.5\% (relative increase) over the next best performing method (PKT-H). At the same time, note that the proposed method seems to lead to overall better results when there is no additional supervision. This is again linked to the existence of critical learning periods. As explained before, forming the appropriate information flow paths requires little to no supervision from the final layers, when the network is randomly initialized (since forming these paths usually change the way the network process information, increasing temporarily the loss related to the final task at hand). 
Similar conclusions can be also drawn from the classification evaluation using the CIFAR-10 dataset. The results are reported in Table~\ref{table:cifar10-classification}. Again, the proposed method leads to a relative increase of about 0.7\% over the next best-performing method.

\begin{table}
	\caption{ Metric Learning Evaluation: STL Distribution Shift}
	\label{table-stl10}
	\centering
	\vspace{0.5em}
	\scriptsize
	\begin{tabular}{l|cccc}
		\textbf{Method} & \textbf{mAP (e)} &  \textbf{mAP (c)}  & \textbf{top-100 (e)} &  \textbf{top-100 (c)} \\
		
		\hline
		
		\hline
		Teacher (ResNet-18) & $57.40$ & $61.20$ & $66.75$ & $69.70$ 
		\\
		Aux. (CNN1-A) &  $44.89$ & $48.48$ & $53.54$ & $56.26$ 
		\\
		Student (CNN1) &  $30.60$ & $33.04$ & $39.08$ & $41.69$ 
		
		\\
		\hline
		Distill &  $33.56$ & $36.23$ & $43.32$ & $46.01$ 
		
		\\
		Hint.  &  $37.11$ & $40.33$ & $46.60$ & $49.46$ 
		
		\\
		MKT &$33.46$ & $35.91$ & $40.65$ & $43.23$ 
		
		\\
		PKT  &$37.22$ & $40.26$ & $44.73$ & $47.98$ 
		
		\\
		\hline
		Hint-H & $35.56$ & $37.85$ & $43.83$ & $46.13$ 
		
		\\
		MKT-H &$33.57$ & $35.23$ & $40.20$ & $42.11$ 
		
		\\
		PKT-H &$37.56$ & $39.77$ & $44.76$ & $47.17$ 
		\\
		Proposed &  $\mathbf{38.11}$ & $\mathbf{40.35}$ & $\mathbf{48.44}$ & $\mathbf{50.57}$
		\\
		
	\end{tabular}

\end{table}

Next, the proposed method was evaluated under a distribution shift setup using the STL-10 dataset (Table~\ref{table-stl10}). For these experiments, the teacher model was trained using the CIFAR-10 dataset, but the KD was conducted using the unlabeled split of the STL dataset. Again, similar results as with the CIFAR-10 dataset are observed, with the proposed method outperforming the rest of the evaluated methods over all the evaluated metrics. Again, it is worth noting that directly transferring the knowledge between all the layers of the network often harms the retrieval precision for the competitive approaches. This behavior is also confirmed using the more challenging CUB-200 data set (Table~\ref{table:cub}), where  the proposed method again outperforms the rest of the evaluated approaches both for the retrieval evaluation, as well as for the classification evaluation. For the latter, a quite large improvement is observed, since the accuracy increases by over 1.5\% over the next best performing method.

Furthermore, we also conducted a HoG~\cite{1467360} cloning experiment, in which the knowledge was transferred from a handcrafted feature extractor to demonstrate the flexibility of the proposed approach. The same strategy as in the previous experiments were used, i.e.,  the knowledge was first transferred to an auxiliary model and then further distilled to the student model. It is worth noting that this setup has several emerging applications, as discussed in a variety of recent works~\cite{passalis2018learning, chen2018distilling}, since it allows for pre-training deep neural networks for domains for which it is difficult to acquire large annotated datasets, as well as providing a straightforward way to exploit the highly optimized deep learning libraries for  embedded devices to provide neural network-based implementations of hand-crafted features. The evaluation results for this setup are reported in Table~\ref{table:handcrafted}, confirming again that the proposed method outperforms the rest of the evaluated methods.

\begin{table}
	\caption{Metric Learning and Classification  Evaluation: CUB-200}
	\centering
	\label{table:cub}
	\vspace{0.5em}
	\scriptsize
	\begin{tabular}{l|ccccc}
		\textbf{Method} & \textbf{mAP (e)} &  \textbf{mAP (c)}  & \textbf{top-10 (e)} &  \textbf{top-10 (c)} & \textbf{Acc.}\\
		
		\hline
		
		\hline
		
		\hline
		Teacher &$63.17$ & $78.17$ & $76.02$ & $81.64$  & $72.16$

		\\
		Aux. &$17.01$ & $18.98$ & $25.77$ & $27.07$  & $32.33$ 
		
		\\
		Student & $15.60$ & $17.24$ & $23.40$ & $24.89$  & $34.08$

		\\
		
		\hline
		
		\hline
		
		\hline

		Distill & $16.40$ & $18.55$ & $24.82$ & $26.57$  & $35.21$

		\\
		Hint.  & $14.34$ & $15.98$ & $22.31$ & $23.41$  & $28.71$

		\\
		MDS & $12.99$ & $13.39$ & $20.60$ & $20.59$  & $30.46$

		\\
		PKT  & $16.36$ & $18.57$ & $24.68$ & $26.70$  & $34.96$

		\\
		\hline
		Hint-H & $13.94$ & $15.37$ & $21.75$ & $22.61$ & $28.34$ 
		
		\\
		MDS-H &  $13.83$ & $15.39$ & $21.27$ & $22.76$ & $32.08$

		\\
		PKT-H & $15.58$ & $17.77$ & $23.50$ & $25.39$ & $33.83$ 
		
		\\
		Proposed & $\mathbf{16.70}$ & $\mathbf{19.01}$ & $\mathbf{25.41}$ & $\mathbf{27.67}$  & $\mathbf{36.95}$   
		
		\\

	\end{tabular}
	
\end{table}

\begin{table}
	\caption{HoG Cloning Network: SUN Dataset}
	\label{table:handcrafted}
	\vspace{0.5em}
	\centering
	\footnotesize
	\begin{tabular}{l|ccccc}
		\textbf{Method} & \textbf{mAP (c)} & \textbf{top-1 (e)} &  \textbf{top-10 (c)} \\
		
		\hline
		
		\hline
		
		\hline
		HoG & ${32.06} \pm {1.20}$ & ${62.55} \pm {1.10}$ & ${47.93} \pm {1.73}$  \\
		Aux. & ${29.69} \pm {2.09}$ & ${55.26} \pm {3.03}$ & ${42.34} \pm {3.71}$  \\

		\hline
		
		\hline
		Hint & ${20.87} \pm {2.13}$ & ${44.14} \pm {4.11}$ & ${31.15} \pm {4.48}$  \\
		MDS  & ${21.65} \pm {2.79}$ & ${43.43} \pm {4.87}$ & ${31.29} \pm {4.24}$  \\
		PKT & ${27.22} \pm {2.60}$ & ${49.90} \pm {3.67}$ & ${36.92} \pm {2.64}$  \\
		Proposed & $\mathbf{27.63 \pm 0.62}$ & $\mathbf{51.18 \pm 1.74}$ & $\mathbf{38.59 \pm 1.00}$  

	\end{tabular}
	
\end{table}

Finally, several ablation studies have been conducted. First, in Fig.~\ref{fig:ablation-critical} we evaluated the effect of using the proposed weighting scheme that takes into account the existence of critical learning periods. The proposed scheme indeed leads to faster convergence over both  single layer KD using the PKT method, as well as over the multi-layer PKT-H method. To validate that the improved results arise from the higher weight given to the intermediate layers over the critical learning period, we used the same decaying scheme for the PKT-H method, but with the initial $\alpha_{init}$ set to 1 instead of 100. Next, we also demonstrated the impact of matching the correct layers in Fig.~\ref{fig:ablation-layers}. Several interesting conclusions can be drawn from the results reported in Fig.~\ref{fig:ablation-layers}. For example, note that over-regularization occurs when transferring the knowledge from a teacher layer that has lower MI with the targets (lower NCC accuracy). On the other hand, using a layer with slightly lower discriminative power (Layer 1 of ResNet-18) can have a slightly positive regularization effect. At the same time, using too discriminative layers (Layer 3 and 4 of ResNet-18) can lead to an early collapse of the representation, harming  the precision of the student.  The accuracy of the student  increases only when the correct layers of the auxiliary teacher are matched to the student (Layers 2 and 3 of CNN-1-A).

Furthermore, we also evaluated the effect of using auxiliary models of different sizes on the precision of the student model trained with the proposed method. The evaluation results are provided in Table~\ref{table:ablation}. Two different student models are used: CNN-1 (15k parameters) and CNN-1-L (6k parameters).  As expected, the auxiliary models that are closer to the complexity of the student lead to improved performance compared both to the more complex and the less complex teachers. That is, when the CNN-1 model is used as student, the CNN-1-A teacher achieves the best results, while when the CNN-1-L is used as student, the weaker CNN-1 teacher achieves the highest precision. Note that as the complexity of the student increases, the efficiency of the KD process declines.

\begin{table}
	\caption{Effect of using auxiliary networks of different sizes (CNN order in term of parameters: CNN-1-H $>$ CNN-1-A $>$ CNN-1 $>$ CNN-1-L)}
	\label{table:ablation}
	\centering
	\vspace{0.2em}
	
	\scriptsize
	\begin{tabular}{l|cccc}
		\textbf{Method} & \textbf{mAP (e)} &  \textbf{mAP (c)}  & \textbf{top-100 (e)} &  \textbf{top-100 (c)} \\
		
		\hline
		CNN1-L $\rightarrow$ CNN1 & $35.03$ & $37.89$ & $46.31$ & $49.27$ \\
		CNN1-A $\rightarrow$ CNN1 & $\mathbf{49.20}$ & $\mathbf{53.06}$ & $\mathbf{61.54}$ & $\mathbf{64.24}$ \\
		CNN1-H $\rightarrow$ CNN1 & $48.82$ & $52.77$ & $61.25$ & $63.99$ \\
		
		\hline
		
		CNN1 $\rightarrow$ CNN1-L & $\mathbf{36.49}$ & $\mathbf{39.25}$ & $\mathbf{48.21}$ & $\mathbf{50.88}$  \\
		CNN-1-A $\rightarrow$ CNN-1-L & $35.72$ & $38.61$ & $47.25$ & $50.13$ \\
		CNN-1-H $\rightarrow$ CNN-1-L & $34.90$ & $37.51$ & $45.83$ & $48.50$ \\
		
	\end{tabular}

\end{table}

\section{Conclusions}
\label{section:conclusions}

In this paper we presented a novel KD method that that works by modeling the information flow through the various layers of the teacher model. The proposed method was able to overcome several limitations of existing KD approaches, especially when used for training very lightweight deep learning models with architectures that differ significantly from the teacher, by  a) designing and training an appropriate auxiliary teacher model, and b) employing a critical-learning aware KD scheme that ensures that critical connections will be formed to effectively mimic the information flow paths of the auxiliary teacher.

\section*{Acknowledgment} This work was supported by the European Union's Horizon 2020 Research and Innovation Program (OpenDR) under Grant 871449. This
publication reflects the authors'' views only. The European Commission is
not responsible for any use that may be made of the information it contains.

\appendix

\section{Appendix}

\subsection{Datasets and Evaluation Setups}
The proposed method was evaluated using four different datasets: the CIFAR-10~\cite{krizhevsky2009learning} dataset, the STL-10~\cite{coates2011analysis} dataset, the CUB-200~\cite{welinder2010caltech} dataset and the SUN Attribute~\cite{patterson2012sun} dataset. For the CIFAR-10, the training split was used for training and transferring the knowledge to the student models, while for the retrieval evaluation the training split was also used to compile the database. Then, the test set was used to query the database and measure the retrieval performance of various representations. For the STL-10 dataset we followed the same setup as for the CIFAR-10, but we also used the provided unlabeded training split for transferring the knowledge to the student models. For the CUB-200 we also followed the same setup, however the experiments were conducted using the first 30 classes of the data, due to the significantly restricted learning capacity of the employed student models (recall that among the objectives of the paper is to evaluate the performance of KD approaches for ultra-lightweight network architectures and heterogeneous KD setups).  Finally,  images from the eight most common categories (for which at least 40 images exist) were used for training and evaluating the methods when the SUN Attribute dataset was employed, since a very small number of
images exist for the rest of the categories.  The 80\% of the extracted images was used for training the networks and building the database, while the rest 20\% was used to query the database. The evaluation process was repeated 5 times and the mean and standard deviation of the evaluated metrics are reported. For the SUN attribute dataset, the knowledge was distilled from a $2 \times 2$ HoG features.

For the CIFAR-10 and STL dataset we used the supplied images without performing any resizing (the original $32\times 32$ images were used). However, the training dataset was augmented by randomly performing horizontal flipping and randomly cropping the images using padding of 4 pixels.  A similar augmentation protocol was used for the CUB-200 dataset. However, the images of the CUB-200 dataset were first resized into $256\times 256$ pixels and then a random crop of $224 \times 224$ pixels was used (a center crop of the same size was used during the evaluation process). Also, random rotation up to $20^{\circ}$ was used when training the models. Finally, the images of the SUN attribute dataset were resized into $128 \times 128$ pixels, before feeding them into the network, following the protocol used in~\cite{passalis2018learning}. 

\begin{figure}
	\begin{center}
		\includegraphics[width=0.99\linewidth]{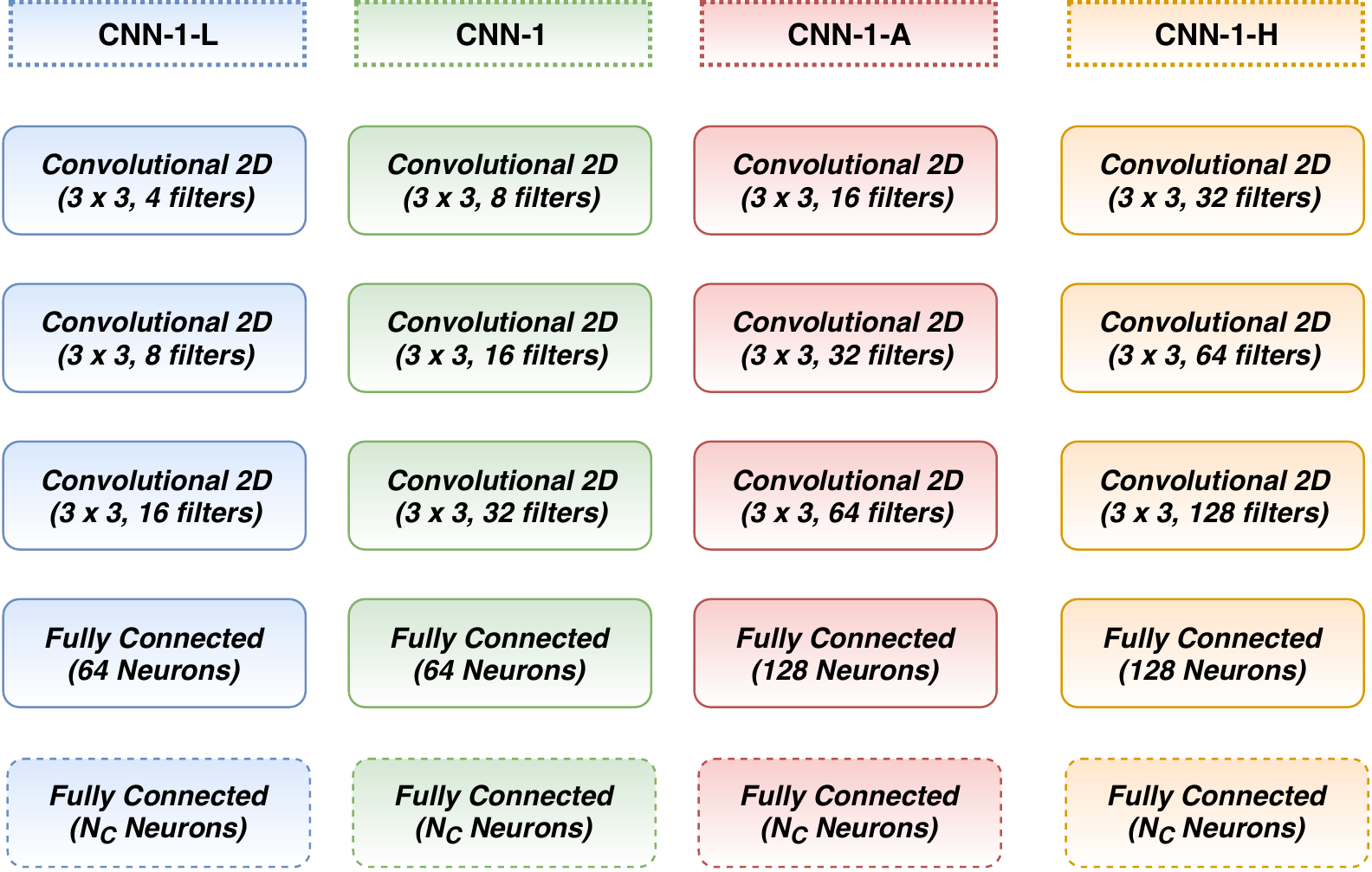}
	\end{center}
	\begin{center}
		\includegraphics[width=0.99\linewidth]{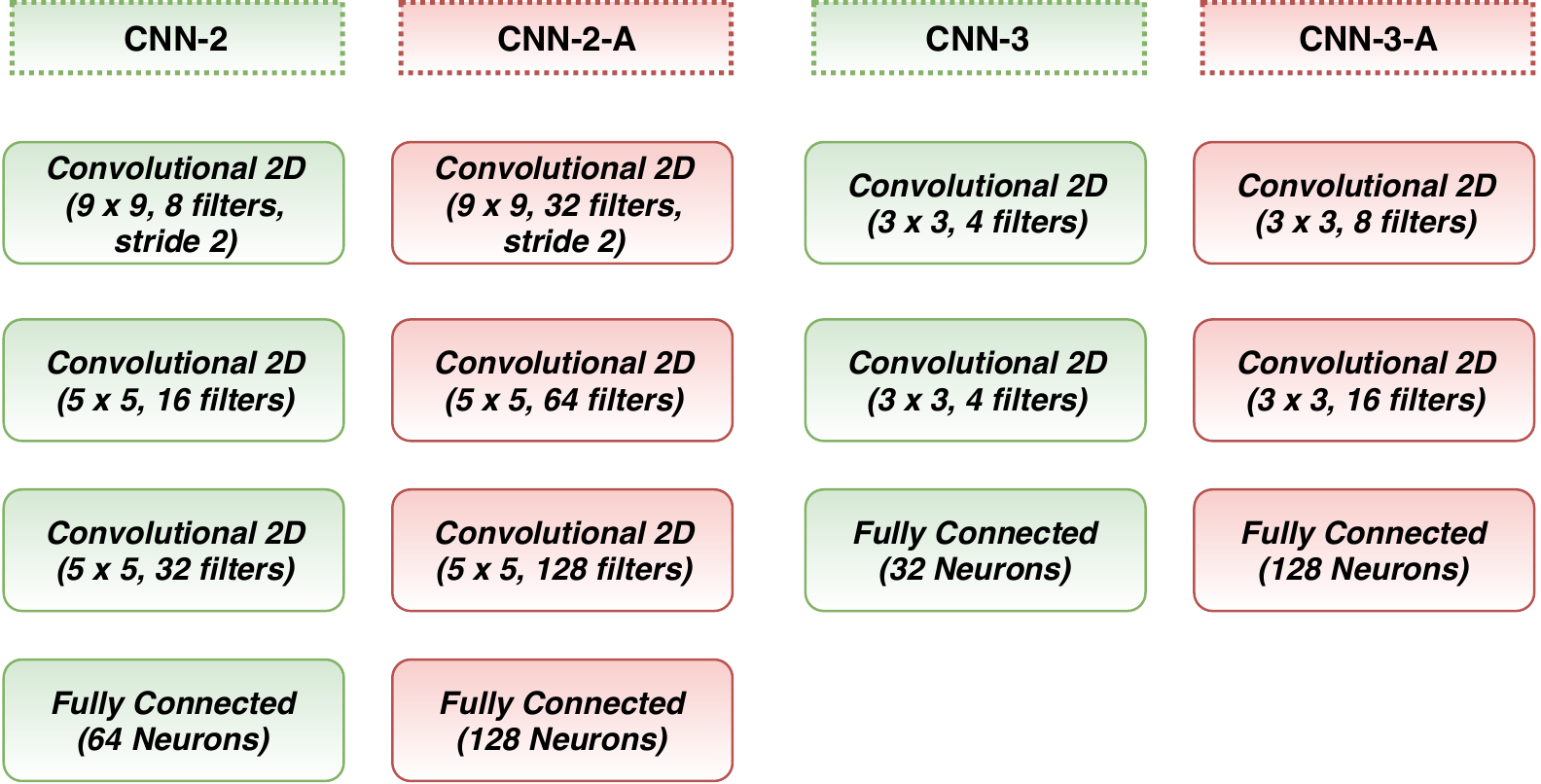}
	\end{center}
	\caption{Network architectures used for the conducted experiments. The green model was used as the student for the conducted experiments (unless otherwise stated), while the red model was used as the auxiliary teacher. For experiments involving classification, an additional  fully connected layer with $N_C$ (number of classes) neurons was added.}
	\label{fig:cnns}
\end{figure}

\subsection{Network Architectures}

The network architectures used for the conducted experiments are shown in Fig.~\ref{fig:cnns}. The CNN-1 family was used for the experiments conducted using the CIFAR-10 and STL dataset, the CNN-2 family was used for the experiments conducted using the CUB-200 dataset, while the CNN-3 family was used for the SUN Attribute dataset. The suffix ``-A'' is used to denote the model that was used as the auxiliary teacher. The auxiliary teacher was trained using the PKT method~\cite{passalis2018learning}, by transferring the knowledge from the penultimate layer of a ResNet-18 teacher (for the CIFAR-10, STL and CUB-200 datasets) or from  handcrafted features (for the SUN Attribute dataset). The ReLU activation function was used for all the layers, while the batch normalization was used after each convolutional layer. 

\subsection{Training Hyper-parameters}

For all the conducted experiments we used the Adam optimizer, with the default training hyper-parameters. For the experiments conducted using the CIFAR-10 dataset the optimization ran for 50 training epochs with a learning rate of 0.001 (batches of 128 samples were used) for all the evaluated methods. For the ablation results reported in Fig.~2 of the main manuscript the optimization ran for 20 epochs. For the STL dataset the optimization ran for 30 training epochs with a learning rate of 0.001 and batch size equal to 128. For the CUB-200 dataset the optimization ran for 100 training epochs, using a learning rate of 0.001 for the first 50 training epochs and 0.0001 for the subsequent 50 training epochs. Also, for the SUN Attribute dataset the optimization ran for 20 training epochs. Furthermore, the decay factor $\gamma$ was set to 0.6 for this dataset, due to the smaller number of training epochs. Finally, note that for the experiments conducted with the contrastive supervision (CIFAR-10) we employed the contrastive loss with the margin set to 1 and the loss was combined with the KD loss after weighting it with 0.1. Also, for the classification experiments reported in Table 2, all the methods were also trained using a supervised classification term (cross-entropy loss). Finally, for all the experiments conducted using the distillation loss, a temperature of $T=2$ was used.

{\small
	\bibliographystyle{ieee_fullname}
	\bibliography{egbib}
}

\end{document}